\documentclass[letterpaper, 10 pt, conference]{ieeeconf}  

\IEEEoverridecommandlockouts                              

\usepackage{amsmath,amssymb,amsfonts}
\usepackage{graphicx}
\usepackage{multirow}
\usepackage{multicol}
\usepackage{makecell}
\usepackage{url}

\usepackage{xcolor}
\usepackage{microtype}
\usepackage{comment}
\usepackage{lineno}
\usepackage{cite}
\usepackage{hyperref}

\usepackage[ruled,vlined,linesnumbered]{algorithm2e}

\DeclareMathOperator*{\argmin}{argmin}

\title{\LARGE \bf
Bilevel Learning for Dual-Quadruped Collaborative Transportation \\
under Kinematic and Anisotropic Velocity Constraints
}

\author{Williard Joshua Jose and Hao Zhang
\thanks{*This work was partially supported by  NSF CAREER Award IIS-2308492 and DARPA Young Faculty Award (YFA) D21AP10114-00.}
\thanks{Williard Joshua Jose and Hao Zhang are with the Human-Centered Robotics Laboratory, University of Massachusetts Amherst, Amherst, MA 01002, USA. Email:
        {\tt\small \{wjose,hao.zhang\}@umass.edu}.}%
}

\begin{document}

\maketitle
\thispagestyle{empty}
\pagestyle{empty}

\begin{abstract}

Multi-robot collaborative transportation is a critical capability that has attracted significant attention over recent years. To reliably transport a kinematically constrained payload, 
a team of robots must closely collaborate and coordinate their individual velocities to achieve the desired payload motion. 
For quadruped robots, a key challenge is caused by their anisotropic velocity limits, 
where forward and backward movement is faster and more stable than lateral motion. 
In order to enable dual-quadruped collaborative transportation and address the above challenges, we propose a novel 
Bilevel Learning for Collaborative Transportation (BLCT) approach.
In the upper-level, BLCT learns a team collaboration policy for the two quadruped robots to move the payload to the goal position, while accounting for the kinematic constraints imposed by their connection to the payload.
In the lower-level, BLCT optimizes velocity controls of each individual robot to closely follow the collaboration policy while satisfying the anisotropic velocity constraints and avoiding obstacles.
Experiments demonstrate that our BLCT approach well enables collaborative transportation in challenging scenarios and outperforms baseline approaches.


More details of this work are provided on the project website: 
\url{https://hcrlab.gitlab.io/project/blct}.
\end{abstract}

\color{black}



\section{INTRODUCTION}

\begin{figure}[t]
\centering
\vspace{6pt}
\includegraphics[width=0.475\textwidth,height=1.8in]{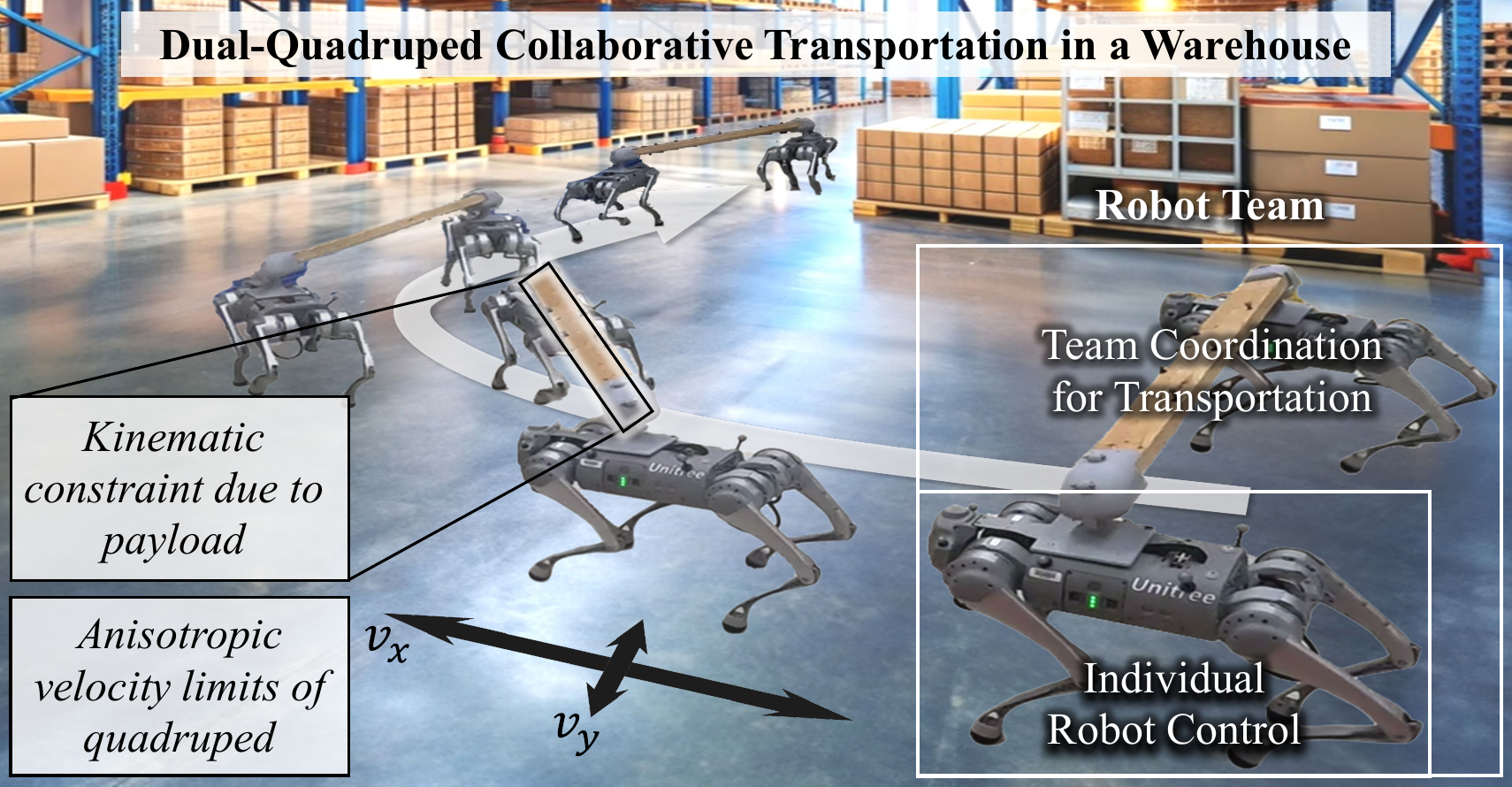}
\vspace{-3pt}
\caption{
A motivating scenario for a pair of quadruped robots collaboratively transport an oversized payload in warehouse environments. 
The robots must not only make team decisions for collaboration
but also generate individual navigational controls,
 while satisfying multiple constraints due to quadruped and payload kinematics, boundary collisions, and anisotropic velocity limits. 
}
\label{fig:motivating_scenario}
\vspace{-6pt}
\end{figure}

Multi-robot systems are becoming an increasingly popular field of robotics that have 
    numerous applications in
    manufacturing \cite{nie_predictive_2024}, 
    search and rescue \cite{queralta_collaborative_2020}, 
    construction \cite{krizmancic_cooperative_2020}, and
    space exploration \cite{farivarnejad_fully_2021}. 
Using multiple smaller, cheaper robots instead of a single large, expensive robot can significantly 
    improve operational efficiency and reduce deployment costs.
Collaborative transportation is an essential multi-robot capability for a team of robots to collaboratively transport payloads to a goal position (Fig. \ref{fig:motivating_scenario}). 
However, controlling robots for collaborative transportation is much more complex and difficult compared to single-robot control,
because individual robots must closely collaborate with their teammates to reliably transport payloads as a team. 

To address multi-robot collaborative transportation, various solutions have been proposed using aerial \cite{tagliabue_robust_2019, wehbeh_distributed_2020}, 
    ground \cite{xia_collaborative_2023, liu_omnidirectional_2023, huang_cooperative_2024}, 
    and legged \cite{vincenti_centralized_2023, yang_collaborative_2022, kim_layered_2023} robots. 
Classic solutions are typically based on motion planning. 
For example, search-based motion planners \cite{le_multi-robot_2019} find navigational plans through graph search, 
but are often compute and memory expensive as the search space grows. 
Sampling-based motion planners \cite{shome_drrt_2020} run faster by sampling and testing possible future states until the goal is reached, but they may generate 
    unnatural motions due to their stochastic nature. 
Potential field-based motion planners \cite{cohen_decentralized_2023} optimize motions to follow attraction and repulsion force vectors, 
but may not always find a valid solution when stuck in local minima.

Recently, several learning-based methods were developed to enable multi-robot collaboration. 
These include works based upon deep neural networks \cite{shi_neural-swarm_2020, riviere_glas_2020} for multi-robot collision avoidance, and graph neural networks \cite{li_graph_2020} for decentralized multi-robot path planning. 
Deep reinforcement learning is also implemented in collaborative transportation to generate multi-robot motion plans \cite{zhang_decentralized_2020} of quadrotors \cite{zhu_learning_2021} and quadrupeds \cite{zhu_cooperative_2023}. 
Moreover, methods are also implemented to plan motion trajectories for human-robot collaborative transportation using deep neural networks \cite{ng_it_2023} and diffusion models \cite{ng_diffusion_2024}. 



Despite the promise in recent learning methods for  multi-robot collaboration, several challenges have not yet been well addressed. 
The first challenge  relates to the potential conflict between team-level decisions and individual robot controls. 
In particular, previous works generally focus only on team-level motion planning while separately implementing individual robot control without having a unified policy integrating the two levels together \cite{liu_omnidirectional_2023, cohen_decentralized_2023}. 
Another challenge arises from the robot team's kinematic constraints, 
where individual robot motions may be constrained (e.g., due to physical attachments to the payload). 
Existing works generally consider only the payload's footprint when dealing with collisions and treat the robots as point masses \cite{zhang_decentralized_2020, riviere_glas_2020, yang_collaborative_2022}, which often fail in space-constrained scenarios. 
Another related challenge is caused by the anisotropic (i.e., {unequal}) velocity limits of the robots. 
For example, although quadruped robots can move in all directions, their motions are faster and more stable 
    when going forward and backward than going sidewards.
Previous learning methods using quadruped robots do not explicitly address the anisotropic velocity constraints\cite{zhu_cooperative_2023, yang_collaborative_2022}.
    
\color{black}

In this paper, 
we introduce a novel \textit{Bilevel Learning for Collaborative Transportation} (BLCT) approach 
to address the problem of Dual-Quadruped Collaborative Transportation (DQCT) where a team of two quadruped robots collaboratively transports a payload to a goal position.
BLCT learns to simultaneously generate team-level collaboration and individual robot controls  
under the unified mathematical framework of bilevel optimization for learning. 
In the upper-level, BLCT learns a collaboration policy for the two quadruped robots to move the payload to the goal position,
while considering the kinematic constraint between the robots due to the connection to the payload.
In the lower-level, 
BLCT optimizes controls of individual robots to 
closely follow the collaboration policy while satisfying 
the anisotropic constraints as well as avoiding collisions with environment obstacles.
To solve the formulated bilevel optimization problem, 
we implement a policy gradient algorithm to simultaneously train the upper-level for team collaboration and the lower-level to optimize individual robot controls under kinematic, collision, and anisotropic velocity constraints. 
\color{black}

The main contribution of this paper is the introduction of our novel BLCT method
to address the DQCT problem.
Two specific novelties include:
\begin{itemize}

    \item We enable a team of quadruped robots to collaboratively transport a payload to a goal position under the kinematic, collision, and anisotropic velocity constraints, 
    which have not been well investigated in the state-of-the-art robotics research yet.
    \item We introduce one of the first mathematical frameworks of constrained bilevel optimization for learning (CBOL) to address the DQCT problem,
    which simultaneously learns a collaboration policy at the upper-level and optimizes controls of individual robots  at the lower-level  to execute the collaboration. 
    
\end{itemize}

\section{RELATED WORK}\label{sec:relatedwork}

\subsection{Multi-Robot Collaborative Transportation}

Collaborative transportation is a problem under multi-robot systems where a team of robots
    must carefully coordinate their actions to collaboratively transport a payload to a 
    goal position \cite{yan_survey_2013, tuci_cooperative_2018, farivarnejad_multirobot_2022}. 
Usually, the payload is either too large (oversized) or too heavy (overweight) to be transported by a single robot. 
The team may have two or more collaborating robots, which may either be homogeneous or heterogeneous. 

Previous works have applied collaborative transportation to teams of unmanned ground vehicles (UGVs) 
    including non-holonomic robots \cite{xia_collaborative_2023, kennel-maushart_payload-aware_2024}
    and omnidirectional robots 
    \cite{liu_omnidirectional_2023, cohen_decentralized_2023, huzaefa_force_2023, huang_cooperative_2024}. 
Similar methods have also been designed for multi-unmanned aerial vehicles (multi-UAVs) 
    based on MPC \cite{tagliabue_robust_2019, wehbeh_distributed_2020} 
    and optimization \cite{wang_cooperative_2018}. 
Recently, collaborative transportation has been implemented on multi-quadruped robot teams, 
    which have higher degrees of freedom and are generally more complex to control. 
Proposed approaches applied MPC 
    \cite{kim_layered_2023, fawcett_distributed_2023, vincenti_centralized_2023} to improve 
    locomotion stability by treating the robot team as a single larger unit, 
    and reinforcement learning \cite{ji_reinforcement_2021} to improve multi-robot coordination. 
However, these works generally rely on having a predetermined trajectory to transport the         
    payload to the goal position generated by a separate motion-planning method. 

Other works also consider the motion planning as part of their problem formulation, such as 
    search-based motion planners \cite{yang_collaborative_2022},  
    sampling-based motion planners \cite{yu_distributed_2022, zhang_hierarchical_2022}, 
    and potential field planners \cite{cohen_decentralized_2023}. 
Recently, reinforcement learning-based planners have been developed to directly output 
    velocity commands for individual robots \cite{zhang_decentralized_2020}. 
However, these works do not directly address individual robot collisions and instead
    only consider collisions caused by the payload. 
When considering omnidirectional robots, these techniques also do not directly model the anisotropic velocity
    limits of the individual robots in the team. 

\subsection{Bilevel Learning for Robotics}

Bilevel learning problems are formulated with an upper-level optimization constrained by the 
    solution space of a nested lower-level optimization problem. 
This framework is increasingly being applied to various problems in computer science and robotics 
    due to its wide applicability \cite{liu_investigating_2021, liu_towards_2021, chen_gradient-based_2023}. 
However, it is still challenging to solve due to usually being a complex and nonconvex
    optimization problem. 

One typical approach is to use the gradient-based bilevel learning, where the gradients from
    the lower-level optimization are propagated to the upper-level to assist in the joint training. 
Gradient-based methods have been applied to robot manipulation \cite{stouraitis_online_2020, zimmermann_multi-level_2020, shirai_robust_2022, wu_learning_2023}, 
    and UAV trajectory optimization \cite{sun_fast_2021, zhu_contact-implicit_2021, chen_simultaneous_2023}. 
However, these methods generally require the gradients to be defined in the problem formulation. 

Another category of approaches solve bilevel optimization by formulating the upper-level problem 
    as a reinforcement learning problem, where the resulting action is used as input to 
    an optmimization problem at the lower-level. 
This has been implemented successfully for neural-architecture search 
    \cite{zoph_neural_2017, zhong_practical_2018}, 
    hyperparameter optimization \cite{dong_dynamical_2021}, 
    and network control \cite{gammelli_graph_2023}. 
However, this formulation has not yet been used in multi-robot collaborative transportation.

\begin{figure*}[t]
\centering
\vspace{6pt}
\includegraphics[width=0.985\textwidth]{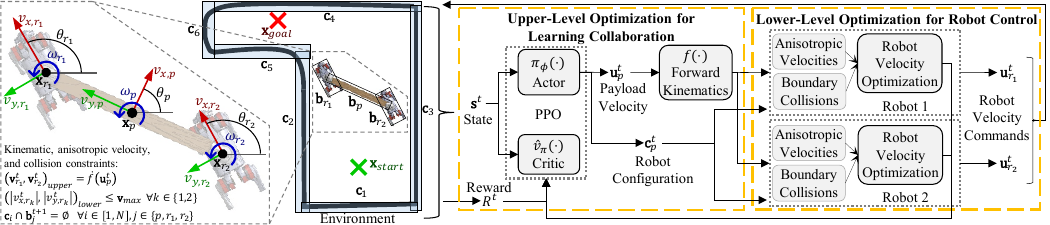}
\vspace{-3pt}
\caption{Overview of our BLCT method.
By formulating collaborative transportation as a bilevel learning problem, 
BLCT uses an upper-level optimization to learn collaborations and a lower-level optimization to generate robot controls, while considering the kinematic, collision, and anisotropic velocity constraints. 
}
\label{fig:formulation}
\end{figure*}

\section{APPROACH}\label{sec:approach}

\textbf{Notation.}
We denote matrices as boldface uppercase letters (e.g., $\mathbf{M}=\{M_{i,j}\}\in\mathbb{R}^{n\times m}$,
an $n\times m$ matrix whose element in the $i$-th row and $j$-th column is $M_{i,j}$),
vectors as boldface lowercase letters (e.g., $\mathbf{v}\in \mathbb{R}^{d}$, a $d$-dimensional vector
whose $i$-th element is ${v}_i$),
and scalars as lowercase letters (e.g., $s$).

\subsection{Problem Formulation}\label{ssec:problem}



We address the problem of dual-quadruped collaborative transportation (DQCT),
    which includes two quadruped robots collaboratively transporting a rigid payload on a flat surface, 
    as illustrated in Fig. \ref{fig:formulation}. 
We use $(\mathbf{x}_p, \theta_p)$ to denote the pose of the payload,
    where $\mathbf{x}_p = (x_p, y_p)$ is its position and $\theta_p$ denotes the orientation of the 
    payload in the global coordinate system.
We use a bounding box, parameterized by $\mathbf{b}_p=(\mathbf{x}_p,\theta_p,\mathbf{d}_p)$, 
    to denote the payload's bounding box that indicates the area occupied by the payload, 
    where $\mathbf{d}_p$ denotes the dimensionality (i.e., height and width) of the bounding box.
We further denote the starting position of the payload as $\mathbf{x}_{start}=(x_{start},y_{start})$ and its goal position as
    $\mathbf{x}_{goal}=(x_{goal},y_{goal})$. 
Similarly, we use $\mathbf{x}_{r_1}$ and $\mathbf{x}_{r_2}$ to denote locations of robots \#1 and \#2, 
    and use $\mathbf{b}_{r_1}$ and $\mathbf{b}_{r_2}$ as parameters to represent their bounding boxes, respectively.

Because each end of the payload is attached to the center of a quadruped robot, this configuration constrains 
    the translation between the quadruped robots while allowing for rotation,
    i.e., the distance between the center positions of the robots remains constant 
    (determined by the length of the payload), but their relative orientation can change.
We represent this \textbf{\textit{kinematic constraint}} between the two robots connected by the payload as a function
    $f: \mathbf{b}_p \mapsto (\mathbf{x}_{r_1}, \mathbf{x}_{r_2})$.


The environment in real-world applications usually contains one or more arbitrarily-shaped boundaries that
     block the path of the payload and quadrupeds, which introduces \textbf{\textit{collision constraints}}.
We propose to approximate these boundaries using a set of $N$ bounding boxes 
    $\{\mathbf{c}_i\}^N_{i=1}$ to fully encapsulate the boundary segments 
    (as illustrated in Fig. \ref{fig:formulation}), 
    where $\mathbf{c}_i=(\mathbf{x}_{c_i},\theta_{c_i},\mathbf{d}_{c_i})$ 
    denotes the bounding box of the $i$-th segment of the boundaries. 
We define the payload to be in collision with the boundaries if 
    $\exists\ i\in[1,N]$ s.t. $\mathbf{c}_i\cap\mathbf{b}_{p}\neq\emptyset$. 
Similarly, we define the same collision conditions for $\mathbf{b}_{r_1}$ and $\mathbf{b}_{r_2}$.

The two quadruped robots are velocity-controlled in a decentralized manner. 
We utilize $\mathbf{u}_{r_1}= (\mathbf{v}_{r_1},\omega_{r_1})$
    to denote the velocity control (i.e., action) of robot \#1,
    where $\mathbf{v}_{r_1} = (v_{x,r_1},v_{y,r_1})$ is the linear velocity 
    and $\omega_{r_1}$ denotes the angular velocity in the robot's coordinate frame.
Similarly, we utilize $\mathbf{u}_{r_2}= (\mathbf{v}_{r_2},\omega_{r_2})$
    to represent the velocity control (i.e., action) of robot \#2.
   
Finally, 
we formulate DQCT as a constrained optimization problem
that obtains a sequence of $\tau$ velocity controls of the two quadrupeds
 $\mathcal{U}=[
        (\mathbf{u}^1_{r_1},\mathbf{u}^1_{r_2}),
        (\mathbf{u}^2_{r_1},\mathbf{u}^2_{r_2}),
        \ldots,
        (\mathbf{u}^{\tau}_{r_1},\mathbf{u}^{\tau}_{r_2})
    ]$
that minimizes the time and distance to transport the payload from its starting position 
 $\mathbf{x}_p^1=\mathbf{x}_{start}$ to its goal position $\mathbf{x}_{goal}$.
 This problem formulation of DQCT can be defined as:
 \begin{align}\label{eq:problem_formulation}
    &\min_{\mathcal{U}}{ 
       \; \tau+\lambda\cdot
        \sum_{t=1}^{\tau}{
            \left(\mathbf{x}^t_p-\mathbf{x}_{goal}\right)^2
        } 
    } \\
    \text{s.t.}\ 
    &(\mathbf{x}^t_{r_1},\mathbf{x}^t_{r_2})=f(\mathbf{b}_p^t)\\
    &\mathbf{c}_i\cap \mathbf{b}^t_j=\emptyset\quad \forall\ 
        i\in[1,N], j\in\{p,r_1,r_2\}
\end{align}
where $\lambda$ denotes a hyperparameter to balance the objectives of minimizing transportation time and 
    minimizing the distance to the goal position. 
The problem formulation addresses DQCT through learning a sequence of $\tau$ velocity controls for the robot team to approach the goal position while considering the     kinematic and collision constraints. 
However, this formulation does not explicitly define the robot team objective
that must be optimized together with individual robot controls. 
In addition, the problem formulation does not consider constraints caused by anisotropic velocity limits of the robots.



\color{blue}

\color{black}

\subsection{Bilevel Learning for Collaborative Transportation}
\label{ssec:dqct_bl}

We introduce a novel \textit{Bilevel Learning for Collaborative Transportation} (BLCT)
    that learns collaboration between the pair of quadruped robots at the upper-level optimization 
    and optimizes controls of individual quadrupeds with anisotropic velocities
    at the lower-level optimization, while considering the kinematic and collision constraints.
An overview of our BLCT approach is illustrated in Fig. \ref{fig:formulation}.



\subsubsection{Upper-Level Optimization for Learning Collaboration}\label{sssec:upper_level}
The goal of the upper-level objective in BLCT is to explicitly learn collaboration 
    between the two robots to collaboratively transport the payload to the goal position with 
    minimum time and distance, as defined in Eq. (\ref{eq:problem_formulation}).
To automatically learn such a collaboration policy, we design the upper-level optimization objective 
    under the framework of reinforcement learning, wherein the robot team carrying the payload 
    learns to transport the payload to the goal position by maximizing task-relevant rewards 
    (e.g., defined by travel time and distance). 

      
Formally, we define the state of the robot team carrying the payload as 
    $\mathbf{s}=(\mathbf{x}_{r_1},\mathbf{x}_{r_2}, \mathbf{b}_p)$, 
    which includes the quadruped positions and payload poses that satisfy the previously defined 
    kinematic constraint $f: \mathbf{b}_p \mapsto (\mathbf{x}_{r_1}, \mathbf{x}_{r_2})$.
We define the action of the quadruped team as $\mathbf{a} = (\mathbf{u}_{r_1},\mathbf{u}_{r_2})$, where $\mathbf{u}_{r}= (\mathbf{v}_{r},\omega_{r})$ denotes the linear and 
    angular velocity control of a robot.
Then, the linear velocities $(\mathbf{v}_{r_1},\mathbf{v}_{r_2})$ of the robots  can be used to 
    compute the velocity $\mathbf{u}_{p}$ of the payload with its ends connected to the center 
    of the quadruped robots.
The same kinematic constraint also applies to constrain the robots' and payload's velocities by the 
    function $\dot{f}: \mathbf{u}_p \mapsto (\mathbf{v}_{r_1}, \mathbf{v}_{r_2})$, 
    which can be used to compute $(\mathbf{v}_{r_1}, \mathbf{v}_{r_2})$ using $\mathbf{u}_p$.


Then, our BLCT's upper-level optimization objective aims to learn a team collaboration policy 
$\pi_{\phi} : \mathbf{s} \mapsto \mathbf{a}$, parameterized by $\phi$,
for the two quadruped robots to collaboratively transport the payload to the goal position,
while considering the kinematic constraints.
This upper-level optimization of our BLCT approach can be mathematically expressed by:
\begin{align}
    &\max_{\pi_{\phi}}\mathbb{E}
        \left[
            \sum_{t=1}^{\tau}\gamma^{t-1} R(\mathbf{s}^t,\mathbf{a}^t)
        \right] \label{eq:upper_level} \\
    \text{s.t.}\ 
    &(\mathbf{v}^t_{r_1},\mathbf{v}^t_{r_2})=\dot{f}(\mathbf{u}_p^t)\label{eq:indiv_vel} \\
    &\mathbf{c}_i\cap \mathbf{b}^t_j=\emptyset\quad \forall\ 
        i\in[1,N], j\in\{p,r_1,r_2\}
\end{align}
where $R$ denotes the reward function, $\gamma\in[0,1]$ is the discount factor,
and $t$ denotes the current time step. 
    



Although quadruped robots can move omnidirectionally, their speed varies by direction of movement, 
which is referred to as {\textit{anisotropic velocities}}. 
Typically, quadrupeds move faster forward along the $x$-axis and slower sideways along the $y$-axis. Moreover, moving forward in the $+x$ direction is usually preferred for intuitive motion around humans and to maintain the use of forward-facing sensors.
The upper-level models anisotropic velocities through introducing a reward function,  
    $R_{vel}=\sum_{i\in\{1,2\}} (v_{x,r_i}-\lvert v_{y,r_i}\rvert)$,
which encourages forward quadruped movement while penalizing backward and lateral motion.
The first term $v_{x,r_i}$ is positive when the $i$-th robot moves forward in the $+x$ direction and negative when moving backward. 
The second term $-\lvert v_{y,r_i}\lvert$ introduces a penalty for lateral movement along the $y$-axis. 
However, as this is a soft constraint, the quadruped is still allowed to move sideways or backward when necessary.

Similarly, the upper-level optimization of BLCT models two additional task-related objectives as reward functions. 
To minimize the travel distance to the goal position, we introduce the reward 
    $R_{dist}=\text{dist}(\mathbf{x}_{start},\mathbf{x}_{goal})-\text{dist}(\mathbf{b}^t_p,\mathbf{x}_{start})$.
To minimize total transportation time, we define a time penalty of $R_{time}=-\tau$.
Incorporating all rewards together, the total reward received by a robot is:
\begin{equation}\label{eq:reward_func}
    R(\mathbf{s},\mathbf{a})=-\tau+\lambda_1 R_{dist}+ \lambda_2 \sum_{t=1}^{\tau} R^t_{vel}
\end{equation}
where $\lambda_1$ and $\lambda_2$ are the hyperparameters to balance the effect of the three rewards.

\color{black}

\subsubsection{Lower-Level Optimization for Robot Control}\label{sssec:lower_level}
The goal of the lower-level optimization in BLCT is to 
optimize the velocity control of each individual quadruped robot under the collision and anisotropic velocity constraints.
Due to the \textbf{\textit{anisotropic velocity constraint}} inherent in quadruped robots, the maximum velocities differ for movements along the $x$-axis and $y$-axis.
We explicitly model this constraint by introducing the maximum anisotropic velocity 
$\mathbf{v}_{max} = (v_{x,max},  v_{y,max})$, where $ v_{x,max} >  v_{y,max} > 0$.
Then, the linear velocities $\mathbf{v}_{r}$ of a quadruped must satisfy the  anisotropic velocity constraint
$\left(\lvert v_{x,r}\rvert,\lvert v_{y,r}\rvert \right) \leq \mathbf{v}_{max}$.

Then, the BLCT's lower-level is formulated as a constrained optimization problem, 
with the objective to guarantee each individual robot's actual velocities (i.e., $\mathbf{v}_{r_1}$ and $\mathbf{v}_{r_2}$)
to closely match the desired velocity controls $(\mathbf{v}_{r_1},\mathbf{v}_{r_2})_{upper}=\dot{f}(\mathbf{u}_p)$ that are generated by $\pi_{\phi}$ at the upper-level for collaborative payload transportation and consider the kinematic constraints. 
At the same time, each individual quadruped must satisfy the anisotropic velocity constraint and the collision constraint for obstacle avoidance.
Formally, the lower-level optimization problem is mathematically defined as:
\begin{align}
    &\min_{\mathbf{u}^t_{r_1},\mathbf{u}^t_{r_2}}
        \|
            (\mathbf{v}^t_{r_1}, \mathbf{v}^t_{r_2})-\dot{f}(\mathbf{u}_p^t)
        \|_2^2
        \label{eq:lower_level} \\
    \text{s.t.}\ 
    &\left(\lvert v_{x,r_k}^t\rvert,\lvert v_{y,r_k}^t\rvert \right) \leq \mathbf{v}_{max}\qquad 
    \forall k \in \{1,2\} \label{eq:lower_velocity}\\
    &\mathbf{c}_i\cap\mathbf{b}^{t+1}_j =\emptyset\qquad \forall\ 
        i\in[1,N], j\in\{p,r_1,r_2\} 
        \label{eq:lower_collision}
\end{align}
The squared $l_2$-loss in the objective function minimizes the difference between the actual velocities of individual robots $(\mathbf{v}_{r_1}, \mathbf{v}_{r_2})$ and the desired velocities computed by $\dot{f}(\mathbf{u}_p^t)$ at the upper-level for the robots to collaboratively transport the payload.
We also explicitly incorporate anisotropic velocities in Eq. (\ref{eq:lower_velocity})
and collision in Eq. (\ref{eq:lower_collision}) as constraints of the lower-level optimization problem.

\subsubsection{Bilevel Learning}\label{sssec:bilevel}
We combined the upper-level and lower-level objectives into our novel BLCT approach under a unified mathematical framework of bilevel optimization for learning, which is mathematically formulated as:
\begin{align}
    &\max_{\pi_{\phi}} \; \mathbb{E}
        \left[
            \sum_{t=1}^\tau\gamma^{t-1} R(\mathbf{s}^t,\mathbf{a}^t)
        \right] \label{eq:final_bilevel_formulation} \\
    \text{s.t.} \notag\\ 
    &\argmin_{\mathbf{u}^t_{r_1},\mathbf{u}^t_{r_2}} \;
         \|
            (\mathbf{v}^t_{r_1}, \mathbf{v}^t_{r_2})_{lower}-\dot{f}(\mathbf{u}_p^t)
        \|_2^2 \label{eq:final_lower_level}\\
            &(\mathbf{v}^t_{r_1},\mathbf{v}^t_{r_2})_{upper}=\dot{f}(\mathbf{u}_p^t)\label{eq:final_kinematic_constraint}\\
          &\left(\lvert v_{x,r_k}^t\rvert,\lvert v_{y,r_k}^t\rvert \right)_{lower} \leq \mathbf{v}_{max}\qquad 
    \forall k \in \{1,2\} \label{eq:final_velocity_constraint} \\
    &\mathbf{c}_i\cap\mathbf{b}^{t+1}_j =\emptyset\qquad \forall\ 
        i\in[1,N], j\in\{p,r_1,r_2\} \label{eq:final_collision_constraint}
\end{align}

BLCT's upper-level in Eq. (\ref{eq:final_bilevel_formulation}) learns a team collaboration policy between the quadrupeds to collaboratively transport the payload,
and the lower-level in Eq. (\ref{eq:final_lower_level}) optimizes controls of individual robots to closely follow the team collaboration policy in a decentralized fashion.
Our BLCT approach also simultaneously considers three constraints. 
Eq. (\ref{eq:final_kinematic_constraint}) represents the kinematic constraint at the upper-level resulting from the connection between the payload and the quadruped robots.
Eq. (\ref{eq:final_velocity_constraint}) represents the anisotropic velocity constraint resulting from the varying capabilities of the quadruped robot moving in different directions.
Eq. (\ref{eq:final_collision_constraint}) models the collision constraint that must be satisfied by both individual robots at the lower-level and the payload carried by the team at the upper-level.

\begin{table*}[ht]
\vspace{6pt}
\caption{Quantitative results of BLCT and comparisons with baseline methods across the left turn, forward bottleneck, \\ and right turn scenarios in Gazebo simulations. 
Planning time, execution time, and total path length are computed based on successful instances. }
\label{table:quantitative}
\centering
\tabcolsep=0.235cm
\begin{tabular}{| c||c|c|c  ||c|c|c||c|c|c||c|c|c|}\hline
\multirow{2}{*}{Method} & \multicolumn{3}{c||}{Success Rate}& \multicolumn{3}{c||}{Planning Time (s)}& \multicolumn{3}{c||}{Execution Time (s)}& \multicolumn{3}{c|}{Total Path Length (m)}\\\cline{2-13}
& Left & Forward & Right  & Left & Forward & Right  & Left & Forward & Right  & Left & Forward &Right  \\
\hline\hline
A* & 69\%& 63\%& 71\%& 0.3922& 0.2356& 0.4259& \textbf{7.3050}& 5.2697& 7.7213& \textbf{4.3736}& \textbf{2.9359}&\textbf{4.5467}\\ \hline
RRT* & 58\% & 68\%& 61\%& 0.7002& 0.2513& 0.4319& 11.5545& 8.6785& 10.6300& 6.4509& 4.8265&6.5601\\ \hline
RL & 34\%& 41\%& 40\%& 0.0962& 0.0782& 0.1033& 9.1226& 8.4317& 9.6377& 4.9399& 5.9102&6.3969\\ \hline
\textbf{BLCT} & \textbf{74\%}& \textbf{72\%}& \textbf{76\%}& \textbf{0.0388}& \textbf{0.0624}& \textbf{0.0610}& 8.1080& \textbf{5.1783}& \textbf{7.6481}& 4.7803& 3.3224&4.9374\\ \hline
\end{tabular}
\end{table*}

To train our BLCT as a bilevel learning method, 
we implement a proximal policy optimization (PPO) \cite{schulman_proximal_2017} algorithm 
that simultaneously learns the team collaboration policy at the upper-level in Eq. (\ref{eq:final_bilevel_formulation})
and optimizes the individual robot velocities at BLCT's lower-level in Eq. (\ref{eq:final_lower_level}),
while satisfying the kinematic, velocity, and collision constraints defined in Eqs. (\ref{eq:final_lower_level})-(\ref{eq:final_collision_constraint}). 
At training time, the upper-level team policy generates desired individual robot velocities using the robot team's velocity kinematics. 
The lower-level policy optimizes the individual controls by scaling the velocity commands
to avoid collisions and respect the anisotropic velocity limits. 
These velocity commands are executed by the robots and the corresponding reward is returned to the upper-level policy to optimize PPO. 
More details on training are included in the supplementary material on the project webpage.


\color{black}




\section{EXPERIMENTS}\label{sec:experiments}

We evaluate our BLCT approach both in Gazebo simulation and a real-robot deployment running Robot Operating System (ROS). 
BLCT is trained and evaluated in simulation running on 8-core Intel Xeon Gold machines with 64GB RAM, while
for real-robot experiments BLCT is executed on onboard Raspberry Pi 4 computers inside the quadruped robots.  
We run our policy at 10Hz for both simulation and real-robots. 
We evaluate on three categories of scenarios in our experiments,
including \textit{left turn}, \textit{forward bottleneck}, and \textit{right turn}. 
These atomic or primary team actions can be chained together to form a longer sequence of motions needed for the quadrupeds to transport an object over longer distances.

We compare our BLCT against three baseline approaches: 
    (1) \textbf{A*} \cite{hart_formal_1968}, a graph-search based motion planning algorithm utilizing discretization of the configuration space,
    (2) \textbf{RRT*} \cite{karaman_incremental_2010}, a sampling-based planning algorithm that progressively builds 
        explores and prunes a tree in the robot's configuration space, and
    (3) \textbf{RL}, our reinforcement learning-only implementation for DQCT without the bilevel formulation and lower-level
        constraints. 
        
We quantitatively evaluate BLCT and compare it against the baseline approaches using four metrics: 
    (1) \textbf{success rate}, defined as the proportion of cases where the collaborative transportation task has been completed, 
    (2) \textbf{planning time}, which is the total time spent computing the robot team's actions, 
    (3) \textbf{execution time}, which is the total time taken by the robot team to transport the payload to the goal position, and
    (4) \textbf{total path length}, which is the total distance traversed by the payload during collaborative transportation. 
Planning time, execution time, and total path length are computed
    based on successful instances.

\begin{figure}[htbp]
\centering 
\vspace{3pt}
\includegraphics[width=0.47\textwidth]{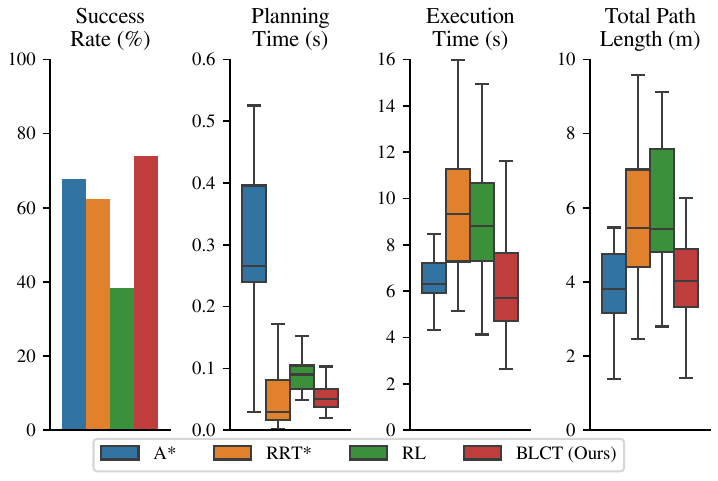}
\caption{Overall performance of BLCT and comparisons with the A*, RRT*, and RL methods. 
BLCT meets or outperforms baseline performance across the four evaluation metrics.}
\label{fig:comparison}
\end{figure}

\begin{figure*}[htbp]
\centering 
\vspace{6pt}
\includegraphics[width=0.985\textwidth]{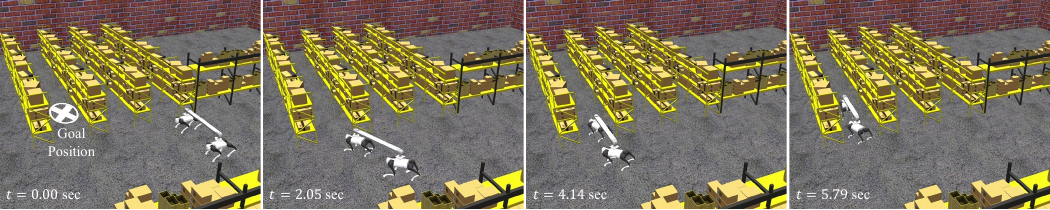}
\caption{Qualitative results obtained by BLCT running on two quadruped robots in Gazebo simulation when the robots collaboratively transport a payload in the right turn scenario. Additional qualitative results for the forward bottleneck and left turn scenarios 
    are on the project website. 
}
\label{fig:qualitative_demo_sim}
\end{figure*}

\begin{figure*}[htbp]
\centering 
\includegraphics[width=0.985\textwidth]{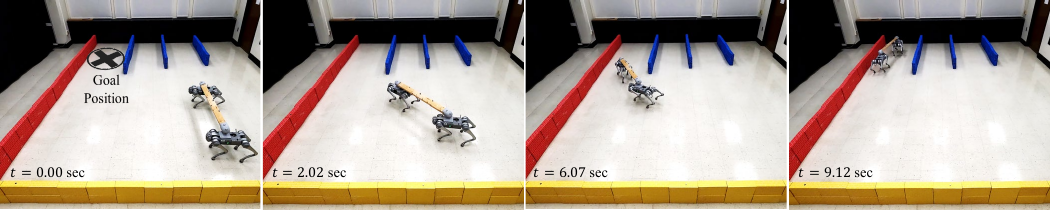}
\caption{Qualitative results obtained by BLCT deployed on two physical Unitree Go1 robots when the robots collaboratively transport a wooden payload in the right turn scenario.
    Additional qualitative results for the forward bottleneck and left turn scenarios are on the project website. 
}
\label{fig:qualitative_demo_real}
\end{figure*}

\subsection{Quantitative Results on Collaborative Transportation}
We evaluate our BLCT approach on the left turn, forward bottleneck, and right turn scenarios in Gazebo. 
The left and right turn scenarios are similar to the visualization in Fig. \ref{fig:formulation}, where the starting position is located
    in a corridor, while the goal position is located in another corridor after the 90-degree turn. 
The forward bottleneck scenario contains a bottleneck where the width of the corridor is narrower. 
We generate 100 unique environments for each scenario with varying corridor dimensions, starting positions, and goal positions. 
In particular, some of the environments may have turns that are too narrow for quadrupeds walking sidewards. 

Our quantitative results are shown in Table \ref{table:quantitative}. 
We observe that BLCT has the highest overall success rate in the scenarios we considered. 
BLCT also has the fastest planning time followed by RL, while A* and RRT* take longer during the start of execution. 
For execution time, BLCT is the fastest in two out of three scenarios, and generally performs as fast as A*. 
BLCT is second to A* for the total path length metric; however, this is expected since A* is designed to be a shortest path algorithm. 

We also analyze the aggregated distribution of the metrics per method in Fig. \ref{fig:comparison}. 
We can observe that the success rate of our BLCT is higher than both A* and RRT*, while RL is significantly lower. 
This illustrates how standard RL does not directly address the DQCT problem and how our bilevel formulation helps learn a robot team
    collaboration policy. 
The planning time of A* (and to a lesser extent, RRT*) varies widely due to differences in the size and complexity of the 
    configuration search space. 
These methods compute the motion plan at the start and may incur some waiting. 
In contrast, BLCT and RL methods compute the next action at regular timesteps and as such have a tighter planning time distribution. 
Additionally, both A* and RRT* generate waypoints which requires a separate controller to generate the motions, while BLCT and RL can both directly output velocity commands.

\subsection{Qualitative Validation using Gazebo Simulations and Real Quadruped Robots}




We employ two Unitree Go1 quadruped robots for running and validating BLCT in simulation. 
The two quadruped robots and a 45-inch payload are modeled in 3D and imported to the Gazebo simulator. 
The payload is attached to the top of the two robots using a ball joint to 
    allow for rotation. 
The quadruped robots are velocity-controlled via ROS using the Champ \cite{champ_github} locomotion controller. 
We obtain robot and boundary poses from Gazebo and ROS directly. 
The qualitative results from our BLCT approach running on Gazebo are demonstrated in Fig. \ref{fig:qualitative_demo_sim}. At time 0.00s, the robots are side-by-side 
    at the starting position.
The right quadruped robot accelerates to start initiating the turn. 
At time 4.14s, both robots start turning towards the corner. 
Finally, at time 5.79s, the robots are at the goal position and stop. 
We can see that now, one robot is in front of the other instead of side-by-side, 
    which highlights the ability of BLCT to reconfigure the robot team's orientation 
    in order to fit tighter spaces. 

Additionally, we utilize two Unitree Go1 quadruped robots to deploy and validate BLCT in a real-world scenario. 
Two ball joints are 3D printed and attached at both ends of a 2kg wooden payload 
    ($45 \times 4 \times 1.5$in), and are then mounted on top of the two quadruped robots. 
The quadruped robots are velocity-controlled using the Go1 SDK and ROS. 
We use an OptiTrack Motion Capture system to determine the ground truth locations
    of the robot team and the boundaries, which are published over the ROS network. 
The qualitative results from our BLCT approach running on the real robots 
    are illustrated in Fig. \ref{fig:qualitative_demo_real}. 
Similar to the simulation, at 0.00s, the real robots stand side-by-side at the starting position. At time 2.02s, the right robot accelerates and moves towards the corner of the turn. 
At time 6.07s, the two robots proceed forward and complete the turn by rounding the corner. 
At time 9.12s, the two robots have arrive in a single file due to the narrow boundaries. 
We also observe that the robots tend to orient themselves towards the direction of motion to execute faster motions due to the anisotropic velocity constraint.

\color{black}

\section{CONCLUSION}
\label{sec:conclusion}

In this paper, we have proposed BLCT as a novel approach for constrained bilevel optimization for learning that addresses the problem of dual-quadruped
    collaborative transportation. 
BLCT learns the robot team collaboration
    policy at the upper-level while simultaneously optimizing individual robot controls  at the lower-level. 
In addition, BLCT explicitly considers
the kinematic constraint caused by the payload that connects the pair of quadruped robots,
the anisotropic velocity constraint of the quadruped robots,
and the obstacle avoidance constraint.
All the constraints are incorporated by BLCT under a unified mathematical framework. 
We evaluated BLCT using Gazebo simulation with multiple environments across three scenarios, 
    and also deployed and validated BLCT on a pair of quadruped robots collaboratively transporting a wooden payload in the real world. 
Our experimental results show that our approach enables the team capability of collaborative transportation, and outperforms baseline methods on multiple metrics. 
Future work can focus on expanding BLCT to perform collaborative transportation on outdoor terrain, 
as well as 
    integrating robot perception to address unstructured environments. 





\section*{ACKNOWLEDGMENT}

The warehouse background image in Fig. \ref{fig:motivating_scenario} is a generative AI image created from Adobe Stock.



\bibliographystyle{IEEEtran}
\bibliography{IEEEabrv,references}

\end{document}